\documentclass[10pt,twocolumn]{article}
\usepackage{graphicx}
\usepackage{amsmath,amssymb,times,cite}
\usepackage{booktabs}
\usepackage{xcolor}

\newcommand{\op}[1]{\operatorname{#1}}
\newcommand{\bm}[1]{\mathbf{#1}} 

\newcommand\T{{\mathpalette\raiseT\intercal}}
\newcommand\raiseT[2]{%
\setbox0\hbox{$#1{#2}$}\raise\dp0\box0}

\title{\Large\textbf{A Graph Encoder-Decoder Network for Unsupervised Anomaly Detection}}
\author{Mahsa Mesgaran and A. Ben Hamza$^*$\\
Concordia Institute for Information Systems Engineering\\
Concordia University, Montreal, QC, Canada\\
$^*$hamza@ciise.concordia.ca}
\date{}

\begin{document}
\maketitle

\begin{abstract}
A key component of many graph neural networks (GNNs) is the pooling operation, which seeks to reduce the size of a graph while preserving important structural information. However, most existing graph pooling strategies rely on an assignment matrix obtained by employing a GNN layer, which is characterized by trainable parameters, often leading to significant computational complexity and a lack of interpretability in the pooling process. In this paper, we propose an unsupervised graph encoder-decoder model to detect abnormal nodes from graphs by learning an anomaly scoring function to rank nodes based on their degree of abnormality. In the encoding stage, we design a novel pooling mechanism, named LCPool, which leverages locality-constrained linear coding for feature encoding to find a cluster assignment matrix by solving a least-squares optimization problem with a locality regularization term. By enforcing locality constraints during the coding process, LCPool is designed to be free from learnable parameters, capable of efficiently handling large graphs, and can effectively generate a coarser graph representation while retaining the most significant structural characteristics of the graph. In the decoding stage, we propose an unpooling operation, called LCUnpool, to reconstruct both the structure and nodal features of the original graph. We conduct empirical evaluations of our method on six benchmark datasets using several evaluation metrics, and the results demonstrate its superiority over state-of-the-art anomaly detection approaches.
\end{abstract}


\section{Introduction}
Graph anomaly detection generally refers to the task of identifying graph nodes that exhibit unusual or unexpected behavior based on their structure and/or feature information. Capturing these abnormal nodes is challenging, primarily because anomalies are rare occurrences and only a very small proportion of the graph nodes might be anomalous. Specific applications of graph anomaly detection include fraud detection, network intrusions, abnormal behavior in social networks, biological systems, communication networks, and financial transactions~\cite{pang2021toward,liu2022dagad}. For example, in social networks, graph anomaly detection can be used to identify fraudulent accounts or suspicious activities.

Detecting anomalies in a graph typically involves identifying nodes that deviate significantly from the normal behavior of the graph, either in terms of their structural characteristics and/or their feature attributes~\cite{Ma21Survey}. However, graphs can often be very large and complex, making it challenging to identify such anomalies. To address this problem, graph pooling can be used to reduce the size of the graph while preserving its important structural features~\cite{ying2018hierarchical,lee2019self,bianchi2020spectral}. The aim is to produce a coarse representation of the graph structure by summarizing the information contained in the nodes of the graph into a fixed-size vector or matrix while preserving the salient features of the graph. By producing a coarse representation of the graph structure, graph pooling can help abstract away irrelevant or noisy details, and focus on the most important structural properties of the graph. Graph pooling methods can be categorized into two main types: global and hierarchical pooling. The former aggregates all of the node features in the graph into a single vector or scalar~\cite{Duvenaud15Molecular,Zhang18SortPool,Zheng21Framelets}. This is typically done using summary statistics (e.g., mean or maximum) to aggregate the features of a set of graph nodes. The resulting global feature can then be used as input to a downstream classifier or regressor. Hierarchical pooling, on the other hand, operates at multiple levels of the graph hierarchy~\cite{ying2018hierarchical,lee2019self,bianchi2020spectral}. The idea is to recursively partition the graph into smaller subgraphs, and then apply pooling at each level to obtain a hierarchy of graph representations. This can be done using clustering-based techniques to group nodes with similar features and represent each group with a single node, resulting in a smaller graph with fewer nodes. More recently, unsupervised methods, which do not require labeled data during training, have been shown effective at addressing anomaly and fault detection problems in various data settings. Tao \textit{et al.}~\cite{Tao2023Bearing} introduce an unsupervised cross-domain diagnosis method to learn fault features specific to the target domain using unlabeled data from the source domain. Song \textit{et al.}~\cite{Song2023DSC} present an adaptive neural finite-time resilient dynamic surface control strategy to overcome unknown control coefficients induced by severe faults and false data injection attacks. Also, Song \textit{et al.}~\cite{Song2023Bit} propose an adaptive fixed-time prescribed performance trajectory tracking controller, incorporating an event-triggered control mechanism, while considering the trade-off between tracking performance and communication cost.

There are several approaches to graph anomaly detection that can be used to identify anomalous nodes in a graph. One common approach is to use unsupervised methods~\cite{perozzi2016scalable,li2017radar,peng2018anomalous,ding2019deep}, which involve identifying anomalous nodes in a graph without using any labeled data. This can be particularly challenging, as there is no ground truth for what constitutes an anomaly. Recently, graph neural networks (GNNs) have achieved state-of-the-art performance in anomaly detection tasks, due largely to their ability to learn efficient representations of nodes that capture their structural or attribute similarity~\cite{Ma21Survey,tang2022rethinking}. GNNs learn node representations by aggregating information from the neighboring nodes of each node in the graph, and hence they are able to capture both the attributes of each node as well as the structure of the graph as a whole. By analyzing both the node attributes and the graph structure simultaneously, GNNs can identify nodes that are significantly different from their neighbors or that exhibit unusual patterns of behavior.

While strong in learning node representations, GNN-based methods are, however, unable to aggregate the information in a hierarchical way~\cite{ying2018hierarchical}. To this end, an effective aggregation function is required to aggregate the information at the node level as well as the entire graph. Such an aggregation strategy is performed using a graph pooling operation, which involves merging nodes or clusters of nodes in the input graph to create a coarser, smaller graph while preserving important structural information of the input graph. Graph pooling can be thought of as a type of downsampling, where the goal is to create a smaller version of the input graph that captures the most salient features of the original graph. Clustering algorithms such as $K$-means or spectral clustering are usually employed to group nodes based on their structural similarities. Hierarchical pooling methods, for instance, reduce the graph size by dropping graph nodes based on a learnable score or by coarsening the graph using a cluster assignment matrix to group nodes into a pre-defined number of clusters. DiffPool~\cite{ying2018hierarchical} is a graph pooling method, which  uses differential graph convolutions to learn an assignment matrix mapping each node to a set of clusters that are used to create a coarser, smaller graph. However, this assignment matrix is dense, and hence it is not easily scalable to large graphs\cite{ranjan2020asap}. SAGPool~\cite{lee2019self} is a graph pooling method, which leverages self-attention networks to learn node scores and construct a smaller sized graph by selecting a subset of nodes based on their scores. However, this technique is unable to preserve node and edge information effectively and they suffer from information loss. Moreover, SAGPool may not be well-suited for handling large graphs with many nodes and edges, as the self-attention mechanism can become computationally expensive for such graphs.

Inspired by locality-constrained linear coding (LLC)~\cite{wang2010locality}, we introduce an unsupervised graph encoder-decoder model for anomaly detection. LLC is a variant of sparse coding that imposes a locality constraint on the weight coefficients. This locality constraint helps capture the local structure of the input signals, leading to a better generalization performance compared to traditional sparse coding. Our proposed model architecture is comprised of two main components: an encoder and a decoder. In the encoding stage, a graph convolutional network encoder is used to generate a latent representation, followed by a graph pooling layer to coarsen the graph. In the decoding stage, an unpooling layer is applied to the coarser graph, followed by a graph deconvolutional network decoder to reconstruct the graph. Our objective is to design a graph pooling operation that is trainable, devoid of learnable parameters, and capable of scaling up to handle large graphs. To this end, we propose a locality-constrained pooling strategy, dubbed LCPool, which generates a coarser graph using a cluster assignment matrix obtained via LLC by solving a constrained least square fitting problem with a locality regularization term. In contrast to sparse coding models, which often rely on computationally intensive optimization algorithms, the objective function used by LLC has an analytical solution~\cite{wang2010locality}, making it perform very fast in practice. Since the deconvolution operation in the decoding stage may introduce undesirable noise into the output graph due to overlapping receptive fields and information loss during downsampling, we employ spectral graph wavelets to enable the preservation of essential details and the removal of unwanted noise, resulting in improved reconstruction of the node feature matrix. The basis functions for the spectral graph wavelets are constructed using the heat kernel, which captures the localized frequency content of a graph signal, allowing for effective feature representation on graph-structured data. To compute these basis functions efficiently, we approximate the heat kernel using a truncated series expansion, thereby providing an efficient way to compute the heat kernel and perform spectral graph wavelet analysis. The main contributions of this work can be summarized as follows:
\begin{itemize}
\item We propose a novel graph encoder-decoder network that effectively learns and encodes underlying patterns and relationships in graph-structured data for anomaly detection.
\item We introduce an effective pooling strategy that focuses on extracting local patterns within graphs, leading to more robust and representative feature encoding.
\item We incorporate a denoising operation into our network architecture using spectral graph wavelets to reduce the impact of noise during the decoding stage with the aim of further improving the quality of the reconstructed graph data.
\item Through extensive experiments on six benchmark datasets, we demonstrate that our proposed method outperforms several state-of-the-art baselines in terms of various evaluation metrics.
\end{itemize}

This paper is structured as follows: In Section 2, we review important relevant work. In Section 3, we introduce a graph encoder-decoder model with a robust pooling strategy for unsupervised anomaly detection. In Section 4, we present experimental results and ablation studies to demonstrate the competitive performance of our approach on six standard benchmarks. Finally, we conclude in Section 5 and identify promising directions for future research.

\section{Related Work}
The basic goal of anomaly detection in attributed graphs is to identify nodes in a graph that exhibit unusual or unexpected behavior based on their attributes or features. Graph pooling, on the other hand, aims to reduce the complexity and size of a graph while preserving its salient features. In this section, we summarize relevant work at the intersection of graph anomaly detection and graph pooling. This integration enables the development of robust approaches that not only detect anomalies effectively but also produce compact graph representations that capture important structural information.

\medskip\noindent\textbf{Graph Anomaly Detection.}\quad The aim of graph anomaly detection is to analyze the topology and attributes of a graph to identify nodes that deviate from the expected patterns. It is a challenging task due to the inherent complexity and structural characteristics of graphs. Graph neural networks, particularly graph convolutional networks (GCNs), have recently become the method of choice for anomaly detection on graphs. These models are predominantly performed in an unsupervised manner due to the cost of acquiring anomaly labels. Li \textit{et al.}~\cite{li2017radar} propose a graph anomaly detection framework that leverages residual analysis to identify deviations between predicted and observed attribute values in the graph. It focuses on detecting attribute-based anomalies in attributed graphs by analyzing the residuals, which are the differences between predicted attribute values and the observed attribute values. Ding \textit{et al.}~\cite{ding2019deep} introduce a GCN-based autoencoder for anomaly detection in attributed graphs by considering both topological structure and nodal attributes. It includes an attributed network encoder designed to capture both network structure and nodal attributes in order to facilitate node embedding representation learning with GCN, a structure reconstruction decoder that reconstructs the original graph topology using the learned node embeddings, and an attribute reconstruction decoder to reconstruct the observed nodal attributes based on the obtained node embeddings. Wang \textit{et al.}~\cite{wang2021one} design a one-class classification method for graph anomaly detection by mapping the training nodes into a hypersphere in the embedding space via graph neural networks. Zhou \textit{et al.}~\cite{zhou2021subtractive} present an abnormality-aware graph neural network, which utilizes a subtractive aggregation technique to represent each node based on its deviation from its neighbors. Nodes that are considered normal and have high confidence are used as labels to train the network in learning a customized hypersphere criterion for identifying anomalies within the attributed graph. Pei \textit{et al.}~\cite{pei2022resgcn} introduce a graph anomaly detection approach that captures the sparsity and nonlinearity present in attributed graphs through the use of GCNs, learns residual information, and employs a residual-based attention mechanism to mitigate the negative impact caused by anomalous nodes. Zhuang \textit{et al.}~\cite{Zhuang2023SIAM} propose a subgraph centralization approach for graph anomaly detection, addressing the weaknesses of existing detectors in terms of computational cost, suboptimal detection accuracy, and lack of explanation for identified anomalies, leading to the development of a graph-centric anomaly detection framework. Duan \textit{et al.}~\cite{Duan2023AAAI} present a multi-view, multi-scale contrastive learning framework with subgraph-subgraph contrast for graph anomaly detection by combining various anomalous information and calculating the anomaly score for each node. However, most of these approaches do not incorporate pooling operations explicitly. Instead, they rely on simple aggregation methods like mean or max pooling to downsample the graph. In contrast, our method introduces a novel pooling strategy based on locality-constrained linear coding, which preserves local structure and captures more informative and discriminative features during the pooling process. Moreover, our method employs a graph encoder-decoder architecture, where the encoder learns high-level features from the graph data, and the decoder reconstructs the original data from these features. This design allows our model to learn more effective and compact representations of the graph, making it better suited for anomaly detection tasks.

\medskip\noindent\textbf{Graph Pooling.}\quad  Graph pooling is a commonly-used operation in graph neural networks, with the aim of producing a compact yet informative representation of the graph structure by summarizing the information contained in the nodes of the graph. By applying a pooling operation, the graph can be transformed into a coarse representation that is easier to analyze or use as input for downstream tasks such as graph anomaly detection. Graph pooling methods can be broadly categorized into two types: global pooling and hierarchical pooling. Global pooling methods summarize the information of all nodes in the graph into a single vector or scalar~\cite{Duvenaud15Molecular,Zhang18SortPool,Zheng21Framelets}, while hierarchical pooling methods recursively apply a pooling operation to the graph, producing a hierarchy of coarser graphs with decreasing numbers of nodes~\cite{ying2018hierarchical,lee2019self,bianchi2020spectral}. On the other hand, spectral clustering pooling techniques consider graph pooling as a cluster assignment task~\cite{bianchi2020spectral}, which categorizes nodes into a set of clusters based on their learned embeddings and constructs the coarser graph based on new nodes using a learned or predefined cluster assignment matrix. Ying \textit{et al.}~\cite{ying2018hierarchical} propose DiffPool, a differentiable pooling strategy that can generate hierarchical representations of graphs by learning a cluster assignment matrix in an end-to-end fashion. This learned assignment matrix contains the probability values of nodes in each layer being assigned to clusters in the next layer generated based on node features and topological structure of the graph. However, DiffPool generates a dense assignment matrix, making it impracticable for large graphs. Moreover, DiffPool can be sensitive to the initial node embeddings used in the clustering process. If the initial embeddings are not well-aligned with the underlying graph structure, it can lead to inaccurate clustering and subsequent pooling, affecting the quality of the learned representations. Other hierarchical pooling methods include SAGPool~\cite{lee2019self} and gPool~\cite{gao2019graph}, which leverage node features and graph topology to learn hierarchical representations. To perform graph pooling, SAGPool selects the most important nodes based on their self-attention scores. The selected nodes are then retained in the pooled representation, while the remaining nodes are discarded. On the other hand, gPool samples nodes according to their scalar projection values using a trainable projection vector, resulting in a coarsened graph. However, self-attention mechanisms and the trainable projection vectors tend to heavily influence the pooling process and are sensitive to the quality of attention or projection vectors. If the attention mechanisms fail to properly capture relevant node relationships or if the trainable vectors are not adequately optimized, it can negatively impact the quality of the pooling operation. Moreover, these pooling operations introduce additional trainable parameters to obtain a coarser graph, thereby increasing the overall complexity of the model. This can lead to a higher number of parameters that need to be learned, resulting in increased memory requirements and computational overhead during training and inference. In addition, these extra trainable parameters in the pooling operations can potentially lead to overfitting, especially when the available training data is limited. In contrast to these pooling operations, our proposed pooling strategy distinguishes itself in several ways. First, it does not rely on learnable parameters, making it more flexible and adaptable. Second, it leverages locality-constrained linear coding to ensure that the local structure of the graph is preserved during the pooling process. Third, it effectively generates a coarser graph representation while preserving the crucial structural attributes that are most relevant to the graph's overall characteristics. By preserving local structure and capturing significant graph characteristics, our pooled strategy enhances the discriminative power of the proposed anomaly detection model, leading to more accurate and reliable results.

\section{Proposed Method}
In this section, we introduce a graph encoder-decoder network for reconstructing both the graph structure and node features. It leverages an encoder-decoder framework to learn and generate a representation of the original graph. By jointly reconstructing both the graph structure and node features, the proposed model can capture and preserve the intricate relationships between nodes and the underlying characteristics of the graph.

\medskip\noindent\textbf{Basic Notions.}\quad An attributed graph is a graph where each node is associated with a set of attributes or features, such as demographic information, transaction history, or social connections. Let $\mathcal{G}=(\mathcal{V},\mathcal{E},\bm{X})$ be an attributed graph, where $\mathcal{V}=\{1,\ldots,N\}$ is the set of $N$ nodes and $\mathcal{E}\subseteq \mathcal{V}\times\mathcal{V}$ is the set of edges, and $\bm{X}=(\bm{x}_{1},...,\bm{x}_{N})^{\T}$ an $N\times F$ feature matrix of node attributes (i.e., $\bm{x}_{i}$ is an $F$-dimensional row vector for node $i$). We denote by $\bm{A}$ an $N\times N$ adjacency matrix whose $(i,j)$-th entry is equal to 1 if $i$ and $j$ are neighboring nodes, and 0 otherwise. We also denote by $\tilde{\bm{A}}=\bm{A}+\bm{I}$ the adjacency matrix with self-added loops, where $\bm{I}$ is the identity matrix.

\medskip\noindent\textbf{Problem Statement.}\quad The goal of unsupervised node anomaly detection in an attributed graph is to identify anomalous nodes in a graph without the use of labeled training data. In other words, there is no available ground truth information that indicates which nodes are anomalous and which ones are not. Given an attributed graph $\mathcal{G}=(\mathcal{V},\mathcal{E},\bm{X})$, the objective of unsupervised node anomaly detection is to learn a scoring function $s:\mathcal{V}\to\mathbb{R}$ that assigns an anomaly score to each node in the graph. Once anomaly scores are computed, the $r$ nodes with the highest anomaly scores are selected based on a user-defined value of $r$. These selected nodes are then identified as anomalies. In other words, nodes with high anomaly scores are considered anomalous, while nodes with lower scores are deemed normal.

\begin{figure*}[!htb]
\centering
\includegraphics[scale=0.65]{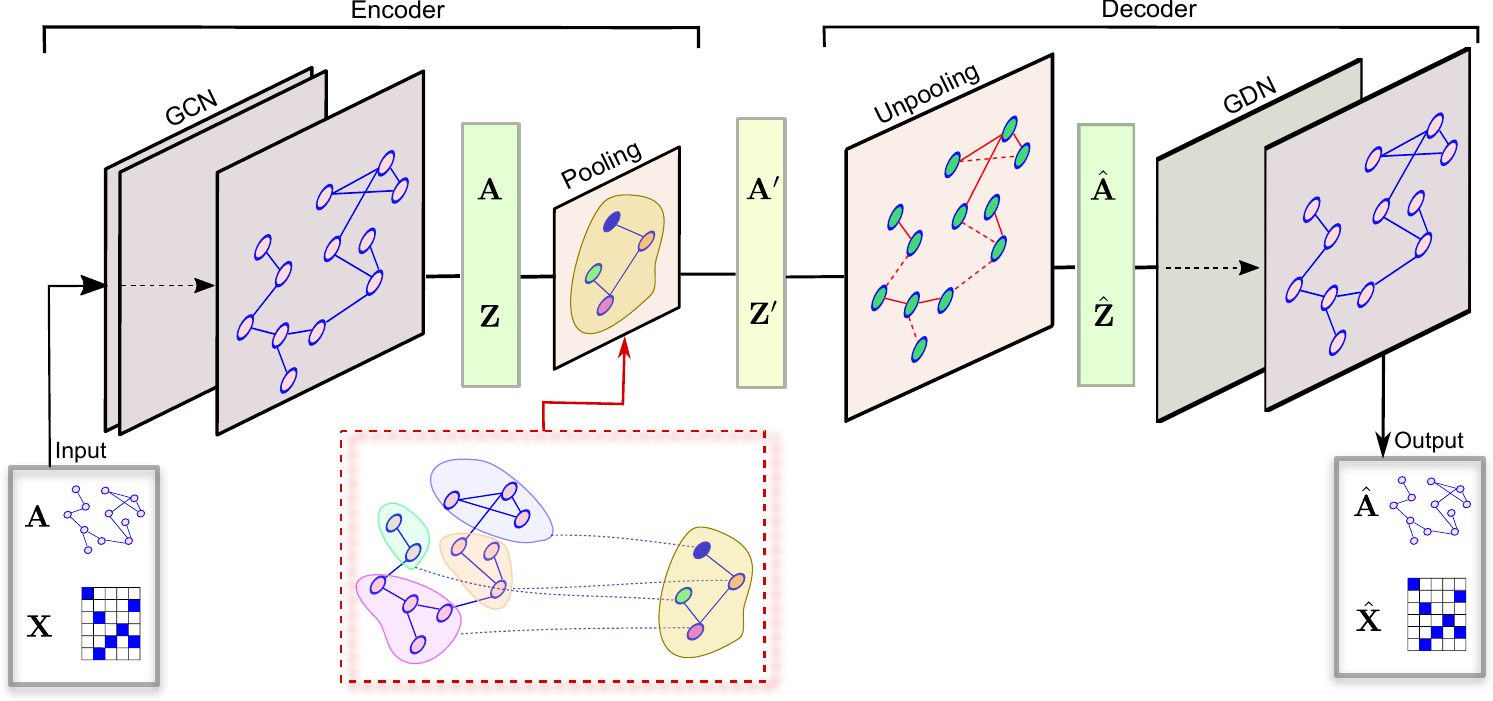}
\caption{Overview of the proposed graph encoder-decoder network architecture. The model consists of two main components: an encoder and a decoder. In the encoding stage, a graph convolutional network (GCN) encoder is used to generate a latent representation, followed by a graph pooling layer to coarsen the graph. In the decoding stage, an unpooling layer is applied to the coarser graph, followed by a graph deconvolutional network (DGN) decoder to reconstruct the graph.}
\label{Fig:GADPool}
\end{figure*}

\medskip\noindent\textbf{Approach Overview.}\quad The overall framework of our proposed approach is shown in Figure~\ref{Fig:GADPool}. The objective is to design a graph encoder-decoder model that can transform an input graph $\mathcal{G}$ into a coarser representation $\mathcal{G}'$, and then reconstruct it back. The proposed model is comprised of two main components: an encoder and a decoder. The encoder consists of a graph convolutional network, which performs convolutions on the graph, and a graph pooling later, which downsamples the graph and extract the most important information from the graph while reducing the number of parameters and computational complexity of the network. The decoder, on the other hand, consists of a graph unpooling layer, which upsamples the representation of the graph, followed by a graph deconvolutional network, which produces an output node feature representation that approximates the input node feature matrix of the original graph.

\subsection{Encoder}
In the encoding stage, a GCN encoder takes as input an adjacency matrix and a node feature matrix, and produces a latent representation of the graph that captures its structural and feature information by performing convolutions on the graph and aggregating information from neighboring nodes. Then, a graph pooling layer is applied to generate a coarser graph representation using a sparse coding based approach, resulting in a coarsened adjacency matrix and a coarsened node embedding matrix.

\medskip\noindent\textbf{GCN Encoder.}\quad The basic idea of a GCN~\cite{Kipf:17} is to learn a set of filters that can extract meaningful features from the graph. This is achieved by aggregating information from neighboring nodes to learn a new representation for each node. The layer-wise propagation rule of GCN is based on the graph convolution operation, which computes the output of a node based on the features of its neighboring nodes as follows:
\begin{equation}
\bm{H}^{(\ell+1)}=\sigma(\tilde{\bm{D}}^{-\frac{1}{2}}\tilde{\bm{A}}\tilde{\bm{D}}^{-\frac{1}{2}}\tilde{\bm{H}}^{(\ell)}\bm{W}^{(\ell)}),\quad \ell=0,\dots,L-1
\label{Eq:IS}
\end{equation}
where $\bm{W}^{(\ell)}$ is a learnable weight matrix, $\bm{H}^{(\ell)}\in\mathbb{R}^{N\times F_{\ell}}$ is the input feature matrix of the $\ell$-th layer with $F_{\ell}$ as embedding dimension, $\tilde{\bm{D}}=\op{diag}(\tilde{\bm{A}}\bm{1})$ is the diagonal degree matrix corresponding to the adjacency matrix with self-added loops, $\bm{1}$ is an $N$-dimensional vector of all ones, and $\sigma(\cdot)$ is an element-wise activation function such as ReLU. The input of the first layer is the initial feature matrix $\bm{H}^{(0)}=\bm{X}$.

The final output node embeddings are given by an $N\times P$ feature matrix $\bm{Z}=\text{GCN}(\bm{A},\bm{X})$ generated at the $L$-layer of the GCN encoder as follows:
\begin{equation}
\bm{Z} = \bm{H}^{(L)}=(\bm{z}_1,\dots,\bm{z}_N)^{\T},
\label{Eq:latent}
\end{equation}
where $P$ is the embedding dimension at the final network layer, and $\bm{z}_i$ is the $i$-th row of $\bm{Z}$ representing the output embedding of node $i$. These learned low-dimensional embeddings capture the structural and semantic similarities of the graph nodes.

\medskip\noindent\textbf{Graph Pooling Layer.}\quad The purpose of a graph pooling layer is to reduce the size and complexity of a graph while preserving its important features and structural characteristics. Given the graph adjacency matrix $\bm{A}\in\mathbb{R}^{N\times N}$ of the input graph $\mathcal{G}$ and the output node embeddings matrix $\bm{Z}\in\mathbb{R}^{N\times P}$ of the GCN encoder, our aim is to design a graph pooling strategy that takes the graph $\mathcal{G}$ as input and produces a coarser graph $\mathcal{G}'$ comprised of $K<N$ nodes, with a weighted adjacency matrix $\bm{A}'\in\mathbb{R}^{K\times K}$ and a node embedding matrix $\bm{Z}'\in\mathbb{R}^{K\times P}$. To generate the coarser graph $\mathcal{G}'$, we use a cluster assignment matrix obtained via locality-constrained linear coding~\cite{wang2010locality}, which is a variant of sparse coding that imposes a locality constraint on the weight coefficients, such that each coefficient is only allowed to depend on nearby basis functions. This constraint is enforced by adding a penalty term to the optimization problem that encourages the weight coefficients to be small for distant basis vectors and large for nearby basis vectors.

\medskip\noindent\textbf{Determining the Assignment Matrix.}\quad  Inspired by locality-constrained linear coding (LLC)~\cite{wang2010locality}, we seek to find an assignment matrix by solving a least square fitting problem with a locality regularization term. Let $\bm{Z} = (\bm{z}_1, \dots, \bm{z}_N)^{\T}$ be the embedding feature matrix obtained by the GCN encoder, where $\bm{z}_i$ is a $P$-dimensional embedding vector for node $i$. We denote by $\bm{V} = (\bm{v}_1,\dots, \bm{v}_K)^{\T}\in\mathbb{R}^{K\times P}$ a codebook (also called vocabulary) constructed via clustering by quantizing the $N$ embedding vectors into $K$ basis vectors. These basis vectors are usually defined as the centers of $K$ clusters obtained via $K$-means clustering on the embedding feature matrix $\bm{Z}$, where $\bm{v}_k$ is a $P$-dimensional vector associated to cluster $k$.

In order to perform the pooling operation, we seek to find a cluster assignment matrix $\bm{U} = (\bm{u}_1,\dots,\bm{u}_N)^{\T}\in\mathbb{R}^{N\times K}$ via LLC, where each code $\bm{u}_i$ is a $K$-dimensional vector obtained by solving the following regularized least-squares problem:
\begin{equation}
\bm{u}_i = \arg \min_{\bm{u}_{i}\bm{1}=1} \|\bm{z}_i - \bm{u}_{i}\bm{V}\|^2 + \lambda \|\bm{d}_i \odot \bm{u}_i\|^2,
\label{Eq:SparseCoding}
\end{equation}
where $\lambda$ is a regularization hyperparameter, $\odot$ denotes the element-wise multiplication, $\bm{d}_{i}$ is a $K$-dimensional vector defined as
\begin{equation}
\bm{d}_{i} = (\exp(\lVert\bm{z}_{i}-\bm{v}_{1}\rVert/\delta),\dots,\exp(\lVert\bm{z}_{i}-\bm{v}_{K}\rVert/\delta)),
\end{equation}
which measures the similarity between the $i$-th embedding and all basis vectors in the codebook, and $\delta$ is a parameter to adjust the weight decay speed for the locality adaptor. Note that the elements of the cluster assignment matrix represent the weights of the basis vectors that are used to reconstruct the embedding feature matrix, while enforcing locality regularization. It should also be noted that the LLC code $\bm{u}_i$ is sparse in the sense that it only comprises a few significant values.

In practice, an approximated LLC is employed for fast encoding by removing the regularization term (i.e., locality constraint) and using the $R$ nearest neighbors of $\bm{z}_i$ as a set of basis vectors, thereby reducing the computational complexity from $\mathcal{O}(K^2)$ to $\mathcal{O}(K+R^2)$, where $K$ is the number of basis vectors in the vocabulary and $R\ll K$. Since the value of $K$ is typically small, the LLC algorithm can be executed quickly in practice.

\noindent Finally, we generate the assignment matrix $\bm{S}$ as follows:
\begin{equation}
\bm{S} = \text{softmax}(\bm{U}),
\label{Eq:AssignmentMatrix}
\end{equation}
where the softmax function is applied row-wise. The $i$-th row of the cluster assignment matrix $\bm{S}\in\mathbb{R}^{N\times K}$ represents the probabilities of node $i$ to be assigned to each of the $K$ clusters, and each column represents a cluster.

\medskip\noindent\textbf{Pooling Strategy with Assignment Matrix.}\quad  The cluster assignment matrix assigns each graph node to a specific cluster, and plays an important role in determining the new representation of the coarser graph produced by the graph pooling operation~\cite{ying2018hierarchical}. The coarsening process aims to reduce the size of a graph by grouping nodes into clusters, while preserving the most important structural features of the graph. Specifically, given the adjacency matrix $\bm{A}$ and node embedding matrix $\bm{Z}$ of the input graph, we define the locality-constrained pooling (LCPool) operator or layer as follows:
\begin{equation}
(\bm{A}',\bm{Z}') = \text{LCPool}(\bm{A},\bm{Z}),
\end{equation}
where
\begin{equation}
\bm{A}'=\bm{S}^{\T}\tilde{\bm{A}}\bm{S}\in\mathbb{R}^{K\times K} \quad\text{and}\quad \bm{Z}'=\bm{S}^{\T}\bm{Z}\in\mathbb{R}^{K\times P}
\end{equation}
are the adjacency matrix of the coarser graph and its new matrix of node embeddings, respectively. Note that $\bm{A}'$ is a weighted adjacency matrix representing the connectivity of the clusters. Each row and column of this coarsened adjacency matrix represents a cluster of nodes, while its $(i,j)$-th element represents the connectivity strength between cluster $i$ and cluster $j$.  Similarly, the $(i,j)$-th element of the new matrix of embeddings $\bm{Z}'$ can be viewed as a weighted sum of the elements of the $j$-th column of $\bm{Z}$, where the weights are given by the corresponding elements of the $i$-th row of $\bm{S}$ (i.e., probabilities of node $i$ to be assigned to each cluster of the $K$ clusters).

\subsection{Decoder}
In the decoding stage, we employ a graph unpooling layer, which upsamples the representation of the graph, followed by a graph deconvolutional network (GDN) decoder, which reconstructs the node feature matrix of the original graph. The aim of the unpooling layer is to upsample the graph by mapping the coarser graph structural and feature representations to finer ones. It basically performs the inverse operation of the pooling layer. Specifically, the unpooling operation attempts to reconstruct the original graph structure and nodal features from the coarser graph representation obtained after pooling. By applying the unpooling operation, the model can reconstruct the finer details of the original graph that may have been lost during the pooling process. This allows for a more accurate representation of the graph's structure and nodal features. Also, the unpooling operation can help provide a clearer understanding of the graph by recovering the original graph structure. This allows for better interpretability and analysis of the graph's properties and characteristics. By reconstructing the original graph, the unpooling operation aims to retain the most significant structural characteristics and nodal features. This can be valuable in applications where preserving important information is crucial, such as anomaly detection. The GDN decoder, on the other hand, is used to decode or reconstruct the original graph from the finer representations generated by the unpooling layer.

\medskip\noindent\textbf{Graph Unpooling Layer.}\quad The purpose of the unpooling layer is to reconstruct the original graph structure from the down-sampled feature maps produced by the pooling layer. One advantage of the coarsened adjacency matrix $\bm{A}'$ is that it preserves the most important structural features of the original graph. To reconstruct the graph structure and nodal features of the original graph, we use a locality-constrained unpooling (LCUnpool) strategy, which takes a coarser graph with features as input and produces a finer representation of the original graph, incorporating both the desired structure and features. We define the LCUnpool operator or layer as follows:
\begin{equation}
(\hat{\bm{A}},\hat{\bm{Z}}) = \text{LCUnpool}(\bm{A}',\bm{Z}'),
\end{equation}
where
\begin{equation}
\hat{\bm{A}}=\bm{S}\tilde{\bm{A}}'\bm{S}^{\T}\in\mathbb{R}^{N\times N} \quad\text{and}\quad \hat{\bm{Z}}= \bm{S}\bm{Z}'\in\mathbb{R}^{N\times P}
\end{equation}
are the unpooled adjacency matrix and the unpooled matrix of node embeddings, respectively. Note that this graph unpooling strategy aggregates information from the node neighborhood by leveraging the normalized coarsened adjacency matrix $\tilde{\bm{A}}'$, which takes into account the relative importance of each node in the graph.

\medskip\noindent\textbf{GDN Decoder.}\quad The deconvolution process can be thought of as an inverse operation to graph convolution, with the aim of recovering the original node features. The key idea behind the GDN decoder is to use a learnable deconvolution operation to reverse the convolutional transformation applied by the GCN encoder. This deconvolution operation needs to take into account the graph structure and ensure that the reconstructed features are consistent with the graph topology. Similar to~\cite{Park2019Symmetric,li2021deconvolutional}, we apply a deconvolution operation by taking the unpooled adjacency matrix $\hat{\bm{A}}$ and node representation $\hat{\bm{Z}}$ as input for the GDN encoder, yielding a reconstructed node feature matrix $\bm{H}\in\mathbb{R}^{N\times F}$ given by
\begin{equation}
\bm{H} = \sigma((\bm{I} + \hat{\bm{L}})\hat{\bm{Z}} \bm{W}),
\end{equation}
where $\hat{\bm{L}}=\bm{I}-\hat{\bm{D}}^{-\frac{1}{2}}\hat{\bm{A}}\hat{\bm{D}}^{-\frac{1}{2}}$ is the normalized Laplacian matrix, $\hat{\bm{D}}=\op{diag}(\hat{\bm{A}}\bm{1})$ is the diagonal degree matrix, and $\bm{W}\in\mathbb{R}^{P\times F}$ is a learnable weight matrix.

The deconvolution operation can be seen as a graph diffusion process that spreads the convolutional features back to their original locations, while taking into account the underlying graph structure. However, the graph convolution of the GCN encoder can be interpreted as a special form of Laplacian smoothing~\cite{Li2018LowPass}, which is a graph filtering operation that can be viewed as a low-pass filter that removes high-frequency noise. Hence, applying the inverse operation of the graph convolution may introduce undesirable noise into the output graph. This issue can be remedied using spectral graph wavelet denoising~\cite{li2021deconvolutional}, a graph signal processing technique that aims to remove noise from a graph signal by leveraging a set of wavelet functions, which are usually defined as a set of filters operating on the graph Laplacian eigenvalues.

The advantage of using spectral graph wavelets for denoising is that it is possible to remove noise that corresponds to high-frequency components of the signal, while preserving the low-frequency components that carry the main information of the signal. Moreover, they provide a flexible and adaptive framework for capturing the underlying structure and smoothness of the graph signal. Specifically, let $\hat{\bm{L}}=\bm{\Phi}\bm{\Lambda}\bm{\Phi}^{\T}$ be an eigendecomposition of the normalized Laplacian matrix, where $\bm{\Phi}$ is a matrix whose columns are the eigenvectors (i.e., graph Fourier basis) and $\bm{\Lambda}=\op{diag}(\lambda_1,\dots,\lambda_N)$ is a diagonal matrix comprised of the corresponding eigenvalues. Spectral graph wavelets have shown to allow localization of graph signals in both spatial and spectral domains~\cite{Hammond:11,Chunyuan:13,donnat2018learning,Emad2007GI,Hamza2007IP,Emad2009ISIVP}. Let $g_{s}(\lambda)=e^{-\lambda s}$ be the transfer function (also called frequency response) of the heat kernel with scaling parameter $s$. The spectral graph wavelet basis $\bm{\Psi}_{s}$ is defined as
\begin{equation}
\bm{\Psi}_{s}=\bm{\Phi}\bm{G}_{s}\bm{\Phi}^{\T},
\end{equation}
where
\begin{equation}
\bm{G}_{s}=g_{s}(\bm{\Lambda})=\op{diag}(g_{s}(\lambda_1),\dots,g_{s}(\lambda_N))
\end{equation}
is a diagonal matrix of transformed eigenvalues via the transfer function. Note that $\bm{\Psi}_{s}$ is also referred to as the heat kernel matrix whose inverse $\bm{\Psi}_{s}^{-1}$ is obtained by simply replacing the scale parameter $s$ with its negative value. The spectral graph wavelet basis and its inverse can also be computed efficiently using polynomial approximations via the Maclaurin series, which can be used to approximate the heat kernel on a graph, by expanding it as a polynomial in the normalized Laplacian matrix, and then truncating the series at a finite order~\cite{li2021deconvolutional}.

Therefore, using the feature representation matrix $\bm{H}$ and both the spectral graph wavelet basis and its inverse, the reconstructed node feature matrix $\hat{\bm{X}}=\text{GDN}(\hat{\bm{A}},\hat{\bm{Z}})$ via the GDN decoder can be obtained as follows:
\begin{equation}
\hat{\bm{X}} = \bm{\Psi}_t \text{ReLU}(\bm{\Psi}_t^{-1} \bm{H} \bm{W}_1) \bm{W}_2,
\end{equation}
where $\bm{W}_{1}\in\mathbb{R}^{F\times P}$ and $\bm{W}_{2}\in\mathbb{R}^{P\times F}$ are learnable weight matrices. The reconstructed node feature matrix aims to provide an approximation of the original node feature matrix. By reconstructing the node feature matrix, we can gain insights into the estimated values of node features, understand the patterns within the graph, and utilize this information for further downstream tasks such as anomaly detection.

\subsection{Model Training}
In order to detect anomalies, we minimize the joint reconstruction loss of the nodal attributes and topological structure, which allows us to learn the reconstruction errors. This loss function is defined as a weighted combination of the structure reconstruction error and the feature reconstruction error
\begin{equation}
\mathcal{L} = (1 - \alpha) \|\bm{A} - \hat{\bm{A}}\|_{F}^2 + \alpha \|\bm{X} - \hat{\bm{X}}\|_{F}^2,
\label{Eq:loss}
\end{equation}
where $\|\cdot\|_F$ denotes the Frobenius norm and $\alpha$ is a weighting hyperparameter that controls the weight or importance given to each error term. The structure reconstruction loss quantifies how well our model approximates the original adjacency matrix, while the feature reconstruction loss measures the quality of the reconstructed node feature representation.

Our proposed graph encoder-decoder model can iteratively approximate the input graph by minimizing the loss function $\mathcal{L}$ to learn the weight matrices. The goal is to adjust the model's parameters such that the loss function is minimized during training. This is typically achieved using a stochastic gradient descent optimizer, which iteratively adjusts the parameters based on the gradient of the loss function with respect to the model parameters. We usually stop training when the performance of the model no longer improves after a certain number of iterations or epochs. The final reconstruction errors are then used to compute the anomaly score of node $i$ as follows:
\begin{equation}
\text{s}_i = (1 - \alpha) \|\bm{a}_i - \hat{\bm{a}}_i\| + \alpha \|\bm{x}_i - \hat{\bm{x}}_i\|,
\end{equation}
which is defined as a weighted sum of a structure error term and a feature error term, where $\bm{a}_i$ and $\bm{x}_i$ are the $i$-th rows of $\bm{A}$ and $\bm{X}$ representing the structure and feature vectors of node $i$, while $\hat{\bm{a}}_i$ and $\hat{\bm{x}}_i$ are the $i$-th rows of $\hat{\bm{A}}$ and $\hat{\bm{X}}$ representing the recovered structure and feature vectors. The nodes are sorted in descending order according to their anomaly scores, and the nodes with the highest anomaly scores are identified as anomalies. Note that a high value of the structure error term implies that the $i$-th node in the graph is more likely to be an anomaly based on the graph structure, while a high value of the feature error term indicates an anomalous node from the feature perspective.

\section{Experiments}
In this section, we present our experimental setup and empirical results. Our aim is to assess the effectiveness and performance of the proposed model in comparison with state-of-the-art methods for graph anomaly detection.

\subsection{Experimental Setup}
\noindent\textbf{Datasets.}\quad We evaluate the performance of our method against state-of-the-art approaches on two groups of standard benchmark datasets:

\smallskip\noindent\textbf{\textsf{\footnotesize Citation networks:}} Cora, Citeseer, Pubmed~\cite{Sen:08}, and ACM~\cite{tang2008arnetminer} are citation network datasets, which are publicly available and consist of scientific publications. In these networks, nodes denote published articles and edges represent the citation relationships between articles. Each node is described by a binary feature vector indicating the absence/presence of the corresponding word from the dictionary.

\smallskip\noindent\textbf{\textsf{\footnotesize Social networks:}} BlogCatalog and Flickr~\cite{tang2009relational} are two typical social network datasets acquired from the blog sharing website, BlogCatalog, and the image hosting and sharing website, Flickr, respectively. In these datasets, nodes represent users of websites and links represent the relationships between users. Social network users typically create personalized content, such as blog posts or photo sharing with tag descriptions, which are considered as attributes of the nodes.

\medskip\noindent We follow the same preprocessing procedure from~\cite{ding2019deep, liu2021anomaly,ding2019interactive}. Since there is no ground-truth of anomalous nodes, it is necessary to artificially introduce synthetic anomalies into the clean attributed networks for the purpose of evaluation. To accomplish this, a collection of anomalies, including both structural and contextual anomalies, are injected into each dataset. Dataset statistics are summarized in Table~\ref{Tab:DataStats}.
\begin{table}[!htb]
\setlength{\tabcolsep}{2.3pt}
\caption{Summary statistics of datasets.}
\small
\medskip
\centering
\begin{tabular}{lrrrr}
\toprule[1pt]
Dataset & Nodes & Edges & Features & Anomalies\\
\midrule[.8pt]
BlogCatalog & 5196 & 171743 & 8189 & 300\\
Flickr & 7575 & 239738 & 12407 & 450\\
ACM & 16484 & 71980 & 8337 & 600\\
Cora & 2708 & 5429 & 1433 & 150\\
Citeseer & 3327 & 4732 & 3703 & 150\\
Pubmed & 19717 & 44338 & 500 & 600\\
\bottomrule[1pt]
\end{tabular}
\label{Tab:DataStats}
\end{table}

\medskip\noindent\textbf{Baselines.}\quad To demonstrate the effectiveness of our method, we include strong baselines for anomaly detection, including local outlier factor (LOF)~\cite{breunig2000lof}, structural clustering algorithm for networks (SCAN)~\cite{xu2007scan}, anomaly mining of entity neighborhoods (AMEN)~\cite{perozzi2016scalable}, residual analysis for anomaly detection in attributed networks (Radar)~\cite{li2017radar}, a joint modeling approach for anomaly detection on attributed networks (ANOMALOUS)~\cite{peng2018anomalous}, deep anomaly detection on attributed networks (Dominant)~\cite{ding2019deep}, deep graph infomax (DGI)~\cite{velickovic2018deep}, contrastive self-supervised learning framework for anomaly detection (CoLA)~\cite{liu2021anomaly}, abnormality-aware graph neural network (AAGNN)~\cite{zhou2021subtractive}, one class graph neural network (OCGNN)~\cite{wang2021one}, Graph Deviation Networks (GDN)~\cite{ding2021few}, DGAE-GAN~\cite{chen2022anomaly}, Residual Graph Convolutional Network (ResGCN)~\cite{pei2022resgcn}. For baselines, we mainly consider methods that are closely related to our proposed approach and/or the ones that are state-of-the-art graph anomaly detection techniques.

\medskip\noindent\textbf{Evaluation Metrics.}\quad To evaluate the performance of our proposed model against the baseline methods, we adopt several evaluation metrics, including AUC, Precision$@K$, Recall$@K$, F1$@K$, and NDCG$@K$. AUC summarizes the information contained in the ROC curve, plotting the true positive rate vs. false positive rate at various thresholds~\cite{Pickup16IJCV,Biasotti16VC}. Larger AUC values indicate better performance at distinguishing between anomalous and normal nodes. Considering the list of nodes sorted based on the anomaly score, Precision$@K$ focuses on the proportion of true anomalous nodes, which are included in the top-$K$ position of ranked nodes. Recall$@K$ measures the proportion of known anomalous nodes selected out of all ground-truth anomalies. F1$@K$is the harmonic mean of Precision and Recall. NDCG$@K$ is a measure of ranking quality, which provides a weighted score that favors rankings where anomalous nodes are ranked closer to the top. These evaluation metrics provide insights into the effectiveness of the method in identifying anomalies and distinguishing them from normal instances.

\medskip\noindent\textbf{Implementation Details.}\quad We implement our model in PyTorch, and train it using Adam~\cite{kingma2014adam} optimizer on the BlogCatalog, Flickr, and ACM datasets for 300 epochs, and on the Cora, Citeseer, and Pubmed datasets for 100 epochs. The learning rate for BlogCatalog, Flicker, ACM, and Cora is set to $10^{-4}$, while the learning rates for Citeseer and Pubmed are set to $10^{-5}$ and $10^{-3}$, respectively. For the GCN encoder, we set the number of hidden layers to 3. The embedding dimension is set to 512 for Cora, Citeseer and Pubmed, and to 218 for BlogCatalog, Flicker and ACM. In all experiments, we set the scaling parameter $s$ to 1 in the wavelet bases. For the approximated LLC, we set the number of neighbors $R$ to 5. A reasonable choice for the weighting hyperparameter $\alpha$ of the loss function is between 0.5 and 0.8 for all datasets. All other hyperparameters and initialization strategies are those suggested by the baselines' authors. We tune hyper-parameters using the validation set, and terminate training if the validation loss does not decrease after 10 consecutive epochs.

\subsection{Anomaly Detection Performance}
We evaluate the anomaly detection performance of our approach against strong baseline methods. Table~\ref{Tab:AUC} reports the AUC scores for our model and baselines on the six benchmark datasets. The AUC scores for the baseline methods on the citation networks are taken from~\cite{ding2019deep} and \cite{liu2021anomaly}. The best results are shown in bold, and the second best results are underlined. As can be seen, our method outperforms the baselines on most datasets with relative improvements of 1.23\%, 0.16\%, 0.76\% and 1.85\% in terms of AUC on BlogCatalog, Flickr, Cora and Citeseer, respectively.

\begin{table}[!htb]
\setlength{\tabcolsep}{1.4pt}
\caption{Test AUC (\%) scores on four citation networks and two social networks. Boldface numbers indicate the best performance,
whereas the underlined numbers indicate the second best performance.}
\small
\medskip
\centering
\begin{tabular}{lcccccc}
\toprule[1pt]
Method & BlogCatalog & Flickr & ACM & Cora & Citeseer & Pubmed\\
\midrule[.8pt]
LOF~\cite{breunig2000lof} & 49.15&48.81& 47.38&-&-&-\\
SCAN~\cite{xu2007scan} & 27.27&26.86&35.99&-&-&-\\
AMEN~\cite{perozzi2016scalable} &66.48&60.47&53.37&62.66&61.54&77.13\\
Radar~\cite{li2017radar} & 71.04&72.86&69.36&65.87&67.09&62.33\\
Anomalous~\cite{peng2018anomalous} &72.81&71.59&71.85&57.70&63.07&73.16\\
Dominant~\cite{ding2019deep} & 78.13&74.90&74.94 &81.55&82.51&80.81\\
DGI~\cite{velickovic2018deep} & 58.27&62.37&62.40 &75.11&82.93&69.62\\
CoLA~\cite{liu2021anomaly} & 78.54&75.13&82.37 &\underline{87.79}&\underline{89.68}&\textbf{95.12}\\
OCGNN~\cite{wang2021one} &55.50 &48.91&50.00 &86.97&85.62&74.72\\
GDN~\cite{ding2021few} &54.24 &52.40&69.15&75.77&78.89&69.15\\
AAGNN~\cite{zhou2021subtractive} & \underline{81.84}&\underline{82.99}&\textbf{85.64}&-&-&-\\
DGAE-GAN~\cite{chen2022anomaly}&81.80 &79.50&83.80&-&-&-\\
ResGCN~\cite{pei2022resgcn} &78.50&78.00&76.80&84.79&76.47&80.79\\
\midrule[.8pt]
\textbf{Ours} & \textbf{82.85} & \textbf{83.12} & \underline{84.69} & \textbf{88.46} & \textbf{91.34} & \underline{92.81}\\
\bottomrule[1pt]
\end{tabular}
\label{Tab:AUC}
\end{table}

In Tables~\ref{Tab:Precision}, \ref{Tab:Recall} and \ref{Tab:F1}, we report the results in terms of Precision$@K$, Recall$@K$ and F1$@K$ scores, respectively, on all datasets for various values of $K$ ranging from 50 to 300. As can be seen, the shallow methods such as LOF, SCAN, AMEN, Radar and ANOMALOUS do not show a competitive anomaly detection performance. This is largely attributed to the fact that their mechanisms have limited capability to detect anomalous nodes in graph-structured data with high-dimensional features and/or complex structures. For instance, both LOF and SCAN yield unsatisfactory results due in large part to the fact that they perform anomaly detection without any knowledge about nodal attributes or graph structure. Among the baselines that consider both attributes and structure, AMEN focuses on finding anomalous connected subgraphs rather than nodes, resulting in poor performance. The residual analysis based models, Radar and Anomalous, show superior performance over the conventional anomaly detection methods (LOF, SCAN and AMEN). However, they can only capture the linear dependency because they are based on matrix factorization. Compared to the other deep learning baselines, our proposed anomaly detection model shows a stronger detection performance and generalization ability.

\begin{table*}[!htb]
\setlength{\tabcolsep}{1.6pt}
\caption{Test Precision$@K$ (\%) scores of our approach and baselines on four citation networks and two social networks. Boldface numbers indicate the best performance,
whereas the underlined numbers indicate the second best performance.}
\small
\medskip
\centering
\begin{tabular}{l|cccc|cccc|cccc|cccc|cccc|cccc}
\toprule[1pt]
&\multicolumn{4}{c|}{BlogCatalog}&\multicolumn{4}{c|}{Flicker}&\multicolumn{4}{c|}{ACM}&\multicolumn{4}{c|}{Cora}&\multicolumn{4}{c|}{Citeseer}&\multicolumn{4}{c}{Pubmed}\\
\midrule
$K$ &50& 100&200&300&50&100&200&300&50&100&200&300&50& 100&200&300&50&100&200&300&50&100&200&300\\
\midrule[.8pt]
LOF~\cite{breunig2000lof} & 30.0& 22.0&18.0&18.3& 42.0& 38.0&27.0&23.7&6.0&6.0&4.5&3.7&-&-&-&-&-&-&-&-&-&-&-&-\\
Radar~\cite{li2017radar} & 66.0&67.0&55.0&41.6& 74.0&70.0&63.5&50.3&5.6&5.8&5.2&4.3& -&- &-&-&- &-&-&-&- &- &-&-\\
Anomalous~\cite{peng2018anomalous} &64.0&65.0&51.5&\underline{41.7}&\underline{79.0}&71.0&65.0&51.0&60.0&57.0&51.0&41.0& -&- &-&-&- &- &-&-&- &- &-&-\\
AMEN~\cite{perozzi2016scalable} &60.0&58.0&49.6&38.3&67.0&64.0&55.0&46.1&52.0&49.0&43.2&36.0&54.6&47.2&29.0&23.0&64.0&44.0&23.0&21.6&56.0&54.1&49.0&45.7\\
Dominant~\cite{ding2019deep} &\underline{76.0}&\underline{71.0}&\underline{59.0}&\textbf{47.0}&77.0&\underline{73.0}&\underline{68.5}&\underline{59.3}&\underline{62.0}&59.0&54.0&\textbf{49.7}&\underline{68.0}&\textbf{55.0}&\underline{36.0}&\underline{27.0}&\underline{76.0}&\underline{51.0}&32.0&25.3&70.0&66.0&\underline{63.0}&\underline{56.0}\\
DGI~\cite{ding2021few} &52.0&51.0&43.6&32.3&59.0&57.7&46.0&45.4&46.0&41.4&38.0&35.4&47.1&39.0&25.3&19.1&54.0&36.3&21.0&17.9&49.0&48.0&44.0&39.5\\
CoLA~\cite{liu2021anomaly} &62.0&58.0&39.5&31.0&60.0&51.0&31.5&26.7&\textbf{88.0}&\textbf{71.0}&\underline{57.5}&46.8&66.0&\underline{54.0}&\textbf{41.5}&\textbf{34.3}&58.0&47.0&\textbf{39.0}&\textbf{31.7}&\textbf{76.0}&\underline{69.0}&58.5&55.7\\
\midrule[.8pt]
Ours&\textbf{80.0}&\textbf{75.0}&\textbf{60.0}&31.7&\textbf{84.0}&\textbf{79.0}&\textbf{71.0}&\textbf{63.3}&\textbf{88.0}&\underline{69.1}&\textbf{58.0}&\underline{47.1}&\textbf{74.0}&\textbf{55.0}&31.0&26.0&\textbf{78.0}&\textbf{59.0}&\underline{32.5}&\underline{27.3}&\underline{75.8}&\textbf{69.2}&\textbf{64.3}&\textbf{57.0}\\
\bottomrule[1pt]
\end{tabular}
\label{Tab:Precision}
\end{table*}

\begin{table*}[!htb]
\setlength{\tabcolsep}{1.6pt}
\caption{Test Recall$@K$ (\%) scores of our approach and baselines on four citation networks and two social networks. Boldface numbers indicate the best performance, whereas the underlined numbers indicate the second best performance.}
\small
\medskip
\centering
\begin{tabular}{l|cccc|cccc|cccc|cccc|cccc|cccc}
\toprule[1pt]
&\multicolumn{4}{c|}{BlogCatalog}&\multicolumn{4}{c|}{Flicker}&\multicolumn{4}{c|}{ACM}&\multicolumn{4}{c|}{Cora}&\multicolumn{4}{c|}{Citeseer}&\multicolumn{4}{c}{Pubmed}\\
\midrule
$K$ &50& 100&200&300&50&100&200&300&50&100&200&300&50& 100&200&300&50&100&200&300&50&100&200&300\\
\midrule[.8pt]
LOF~\cite{breunig2000lof} & 5.0& 7.3&12.0&18.3& 4.7& 8.4&12.0&15.8&0.5&1.0&1.5&1.8&- &- &-&-&-&- &-&-&-&-&-&-\\
Radar~\cite{li2017radar} & 11.0&22.3&36.7&41.6& 8.2&15.6&28.2&33.6&4.7&9.7&17.3&21.5&- &- &-&-&- &- &-&-&- &- &-&-\\
Anomalous~\cite{peng2018anomalous} &10.7&21.7&34.3&41.7&\underline{8.7}&15.8&28.9&34.0&5.0&9.5&17.0&20.5&- &- &-&-&- &- &-&-&- &- &-&-\\
AMEN~\cite{perozzi2016scalable} &9.7&19.6&32.4&38.9&7.2&14.3&26.9&32.1&4.5&7.9&15.9&17.1&20.9&31.4&42.4&47.6&21.6&30.1&37.7&44.7&4.4&8.1&15.5&22.2\\
Dominant~\cite{ding2019deep}  &\underline{12.7}&\underline{23.7}&\underline{39.3}&\underline{47.0}&8.4&\underline{16.2}&\underline{30.4}&\underline{39.6}&5.2&9.8&18.0&\underline{24.8}&\underline{23.7}&\underline{36.7}&48.0&54.0&\underline{25.3}&\underline{34.0}&42.7&50.7&\textbf{28.3}&11.0&\underline{21.0}&\underline{28.0}\\
DGI~\cite{ding2021few} &8.4&17.1&28.2&34.1&7.3&13.0&24.4&29.3&4.3&8.4&13.7&16.5&18.1&27.1&31.9&35.8&17.2&24.1&30.3&35.8&3.5&6.6&12.1&17.7\\
CoLA~\cite{liu2021anomaly} &10.4&19.5&26.5&31.2&6.7&11.5&14.1&18.0&\underline{7.3}&\underline{11.9}&\underline{19.3}&23.4&22.0&36.0&\underline{55.3}&\underline{68.7}&19.3&31.3&\underline{52.0}&\underline{63.3}&6.3&\underline{11.5}&19.5&27.9\\
\midrule[.8pt]
Ours&\textbf{24.8}&\textbf{25.2}&\textbf{40.3}&\textbf{60.9}&\textbf{18.2}&\textbf{20.6}&\textbf{41.3}&\textbf{48.8}&\textbf{9.8}&\textbf{18.5}&\textbf{21.0}&\textbf{28.3}&\textbf{42.5}&\textbf{51.7}&\textbf{71.3}&\textbf{88.5}&\textbf{31.0}&\textbf{52.2}&\textbf{57.5}&\textbf{70.8}&\underline{11.8}&\textbf{15.8}&\textbf{27.8}&\textbf{30.8}\\
\bottomrule[1pt]
\end{tabular}
\label{Tab:Recall}
\end{table*}

\begin{table*}[!htb]
\setlength{\tabcolsep}{1.6pt}
\caption{Test F1$@K$ (\%) scores of our approach and baselines on four citation networks and two social networks. Boldface numbers indicate the best performance, whereas the underlined numbers indicate the second best performance.}
\small
\medskip
\centering
\begin{tabular}{l|cccc|cccc|cccc|cccc|cccc|cccc}
\toprule[1pt]
&\multicolumn{4}{c|}{BlogCatalog}&\multicolumn{4}{c|}{Flicker}&\multicolumn{4}{c|}{ACM}&\multicolumn{4}{c|}{Cora}&\multicolumn{4}{c|}{Citeseer}&\multicolumn{4}{c}{Pubmed}\\
\midrule
$K$ &50& 100&200&300&50&100&200&300&50&100&200&300&50& 100&200&300&50&100&200&300&50&100&200&300\\
\midrule[.8pt]
LOF~\cite{breunig2000lof} &8.6&10.9&14.4&18.3&8.4&13.7&16.6&18.9&0.9&1.7&2.2&2.4&-&-&-&-&-&-&-&-&-&-&-&-\\
Radar~\cite{li2017radar} &18.8&33.4&44.0&41.6&14.7&25.5&39.0&40.3&5.1&7.2&7.9&7.2&-&-&-&-&-&-&-&-&-&-&-&-\\
Anomalous~\cite{peng2018anomalous} &18.3&32.5&41.2&41.7&\underline{15.7}&25.8&40.0&40.8&9.2&16.3&25.5&27.3&-&-&-&-&-&-&-&-&-&-&-&-\\
AMEN~\cite{perozzi2016scalable} &16.7&29.3&39.2&38.6&13.0&23.4&36.1&37.8&8.3&13.6&23.2&23.2&30.2&37.7&34.4&31.0&32.3&35.7&28.6&29.1&8.2&14.1&23.6&29.9\\
Dominant~\cite{ding2019deep} &\underline{21.8}&\underline{35.5}&\underline{47.2}&\textbf{47.0}&15.1&\underline{26.5}&\underline{42.1}&\underline{47.5}&9.6&16.8&27.0&\underline{33.1}&\underline{35.1}&\underline{44.0}&41.1&36.0&\underline{38.0}&\underline{40.8}&36.6&33.8&\underline{40.3}&18.9&\underline{31.5}&\underline{37.3}\\
DGI~\cite{ding2021few} &14.5&25.6&34.2&33.2&13.0&21.2&31.9&35.6&7.9&14.0&20.1&22.5&26.2&32.0&28.2&24.9&26.1&29.0&24.8&23.9&6.5&11.6&19.0&24.4\\
CoLA~\cite{liu2021anomaly} &17.8&29.2&31.7&31.1&12.1&18.8&19.5&21.5&\underline{13.5}&\underline{20.4}&\underline{28.9}&31.2&33.0&43.2&\textbf{47.4}&\textbf{45.8}&29.0&37.6&\textbf{44.6}&\textbf{42.2}&11.6&\underline{19.7}&29.3&37.2\\
\midrule[.8pt]
Ours&\textbf{37.9}&\textbf{37.7}&\textbf{48.2}&\underline{41.7}&\textbf{29.9}&\textbf{32.7}&\textbf{52.2}&\textbf{55.1}&\textbf{17.6}&\textbf{29.2}&\textbf{30.8}&\textbf{35.4}&\textbf{54.0}&\textbf{53.3}&\underline{43.2}&\underline{40.2}&\textbf{44.4}&\textbf{55.4}&\underline{41.5}&\underline{39.4}&\textbf{20.4}&\textbf{25.7}&\textbf{38.8}&\textbf{40.0}\\
\bottomrule[1pt]
\end{tabular}
\label{Tab:F1}
\end{table*}

\medskip\noindent\textbf{Model Efficiency.}\quad The training time of our model depends on the complexity of the graph and the size of the dataset, as well as the number of layers and learnable parameters in the model. Applying LCPool in the proposed method does introduce additional computation to compute the assignment matrix. However, in practice, we observed that LCPool does not significantly increase the running time of the model. This is because each LCPool layer effectively reduces the size of the graph by creating a coarser representation of the graph, which leads to a speed-up in the subsequent graph convolution operation in the next layer. The reduction in graph size achieved by LCPool means that the graph convolutional operation in the subsequent layers processes fewer nodes and edges, resulting in a more efficient computation. As a result, any additional overhead from the computation of the assignment matrix is offset by the reduced computation in the subsequent layers. Moreover, LCPool's locality-constrained linear coding mechanism efficiently captures local patterns and generates more compact and informative embeddings, further contributing to the efficiency of the overall model. In addition, once our model is trained, the inference or anomaly detection time for a given graph is relatively fast. The model only needs to perform a forward pass through the encoder and decoder to compute the reconstruction loss and anomaly scores.

\subsection{Parameter Sensitivity Analysis}
\begin{figure}[!htb]
\setlength{\tabcolsep}{-.5em}
\centering
\begin{tabular}{cc}
\includegraphics[scale=.27]{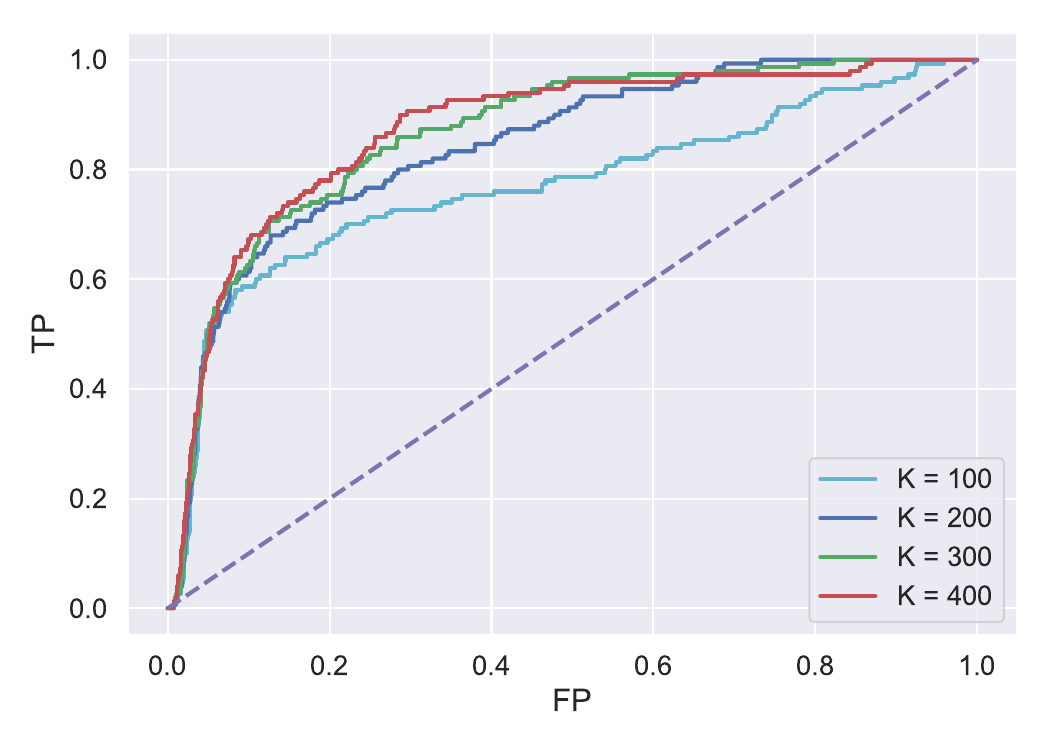}& \includegraphics[scale=.27]{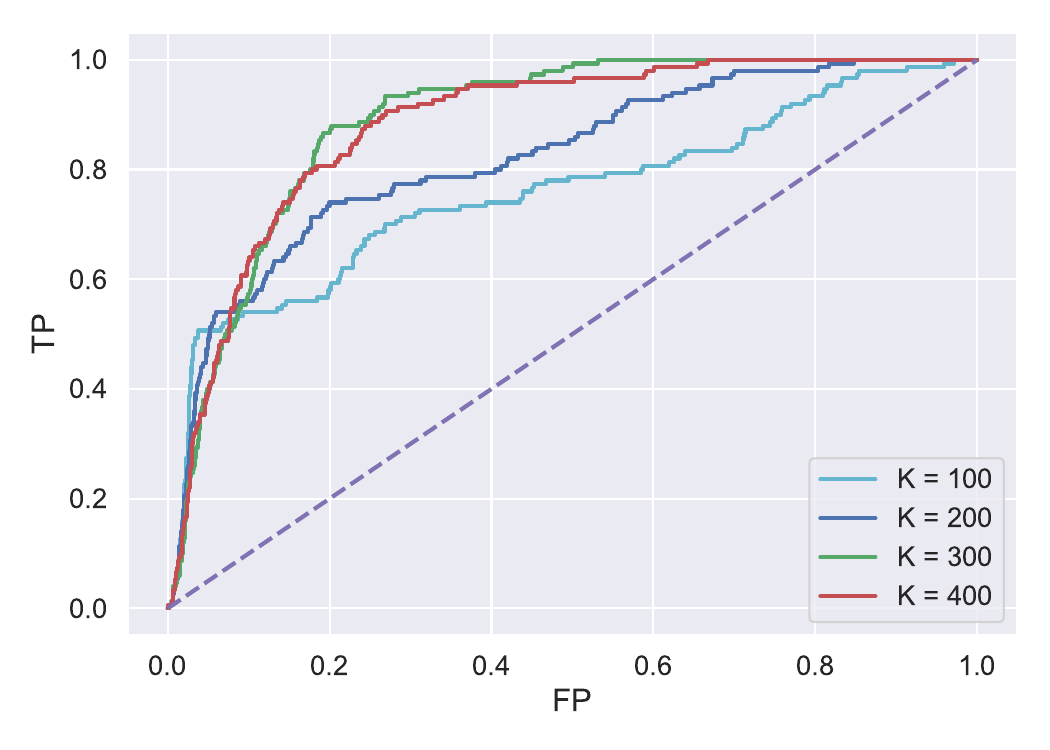} \\
Cora & Citeseer\\
\includegraphics[scale=.27]{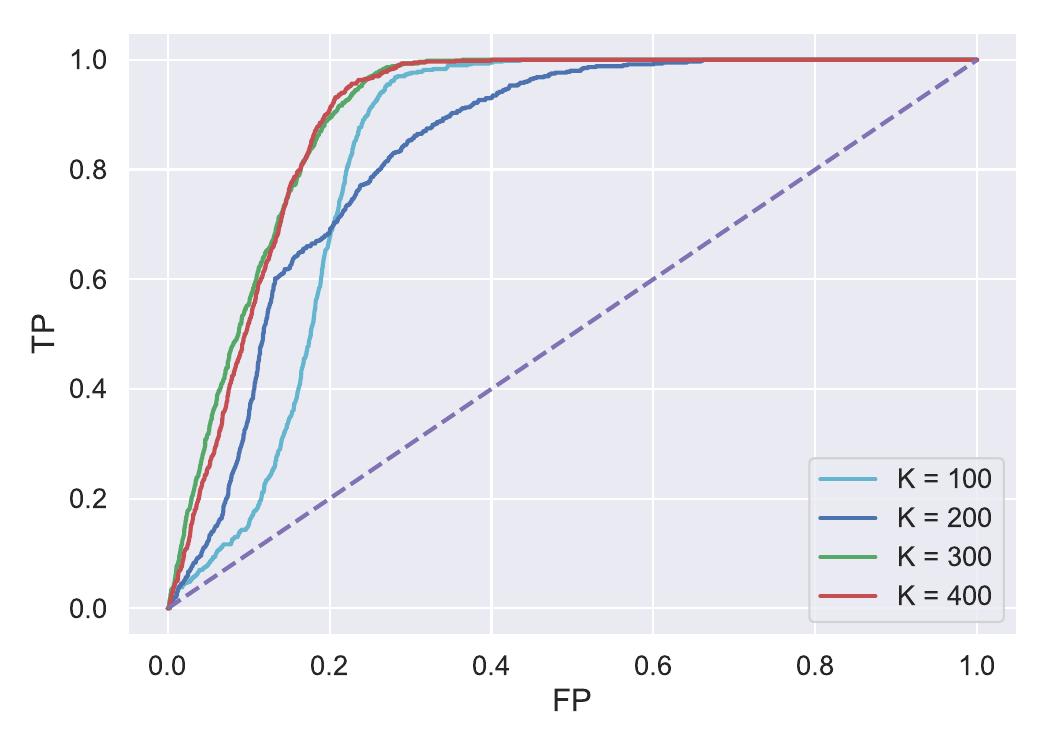} & \includegraphics[scale=.27]{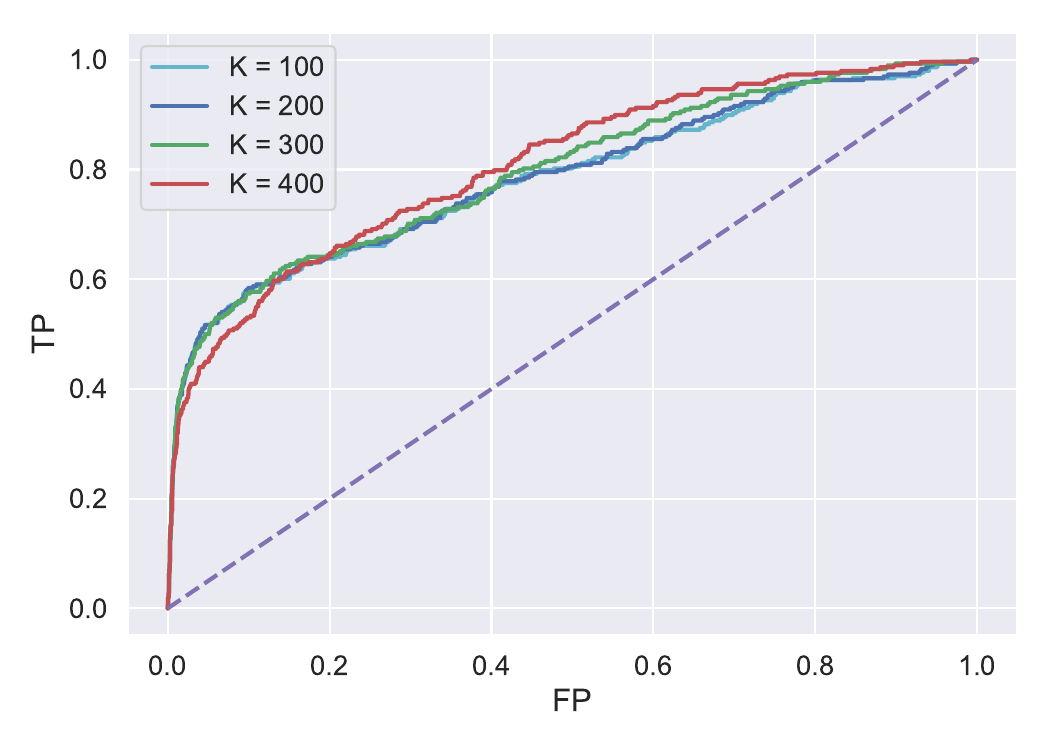} \\
Pubmed & BlogCatalog\\
\includegraphics[scale=.27]{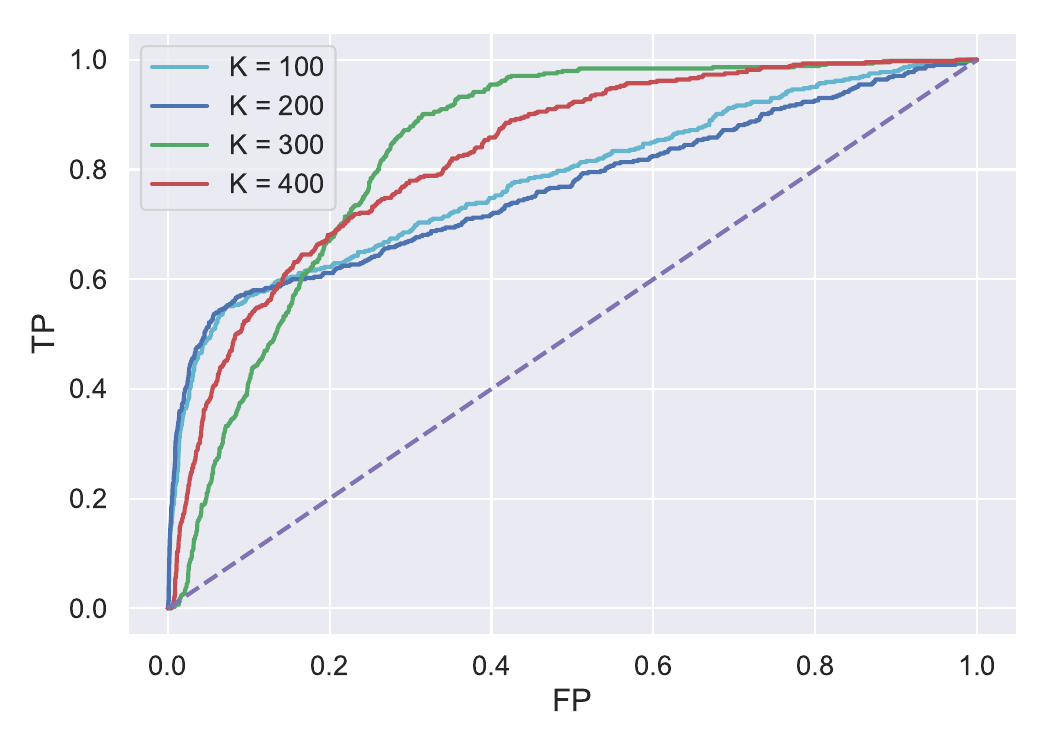}& \includegraphics[scale=.27]{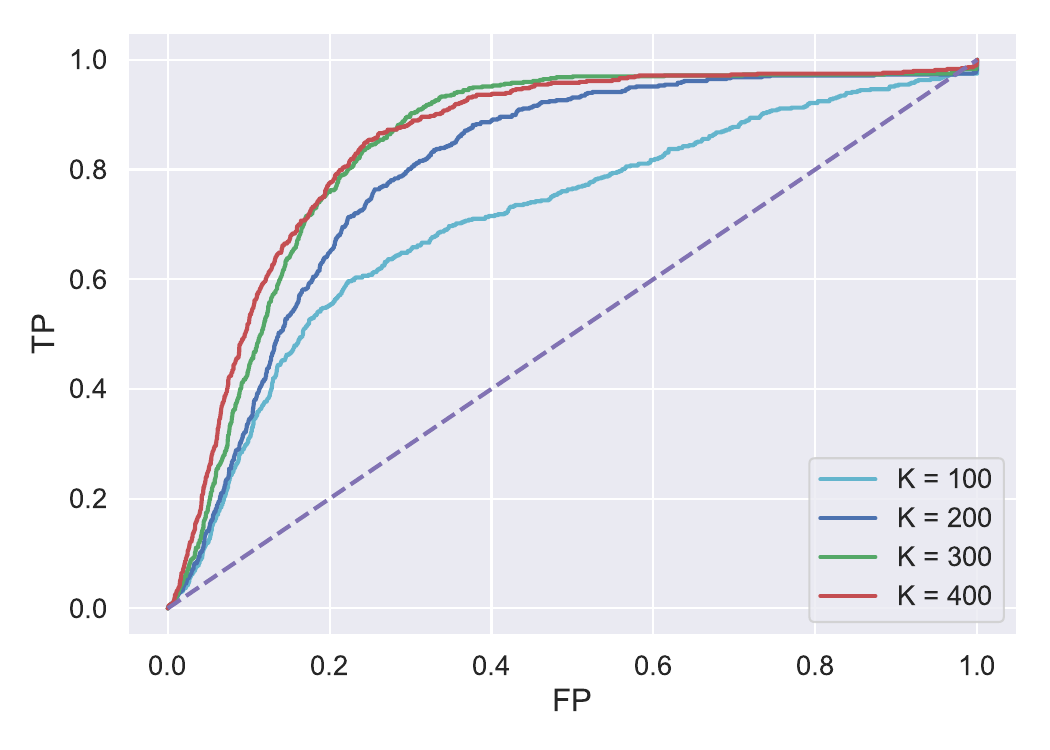} \\
Flickr & ACM
\end{tabular}
\caption{Effect of hyperparameter $K$ on anomaly detection performance of our model using ROC curves.}
\label{Fig:k}
\end{figure}

The hyperparameter $K$, which is the new dimension of the embedded graph after applying graph pooling, plays an important role in the anomaly detection performance of the proposed model. We conduct a sensitivity analysis to investigate how the performance of our approach changes as we vary this embedding dimension. In Figure~\ref{Fig:k}, we analyze the effect of this hyperparameter by plotting the ROC curves for our model on all datasets, where $K$ varies in the set $\{100, 200, 300, 400\}$. We can see that our model generally benefits from relatively larger values of $K$. For all datasets, our model achieves good performance when $K = 400$.

\begin{figure}[!htb]
\setlength{\tabcolsep}{0em}
\centering
\begin{tabular}{c}
\includegraphics[scale=.5]{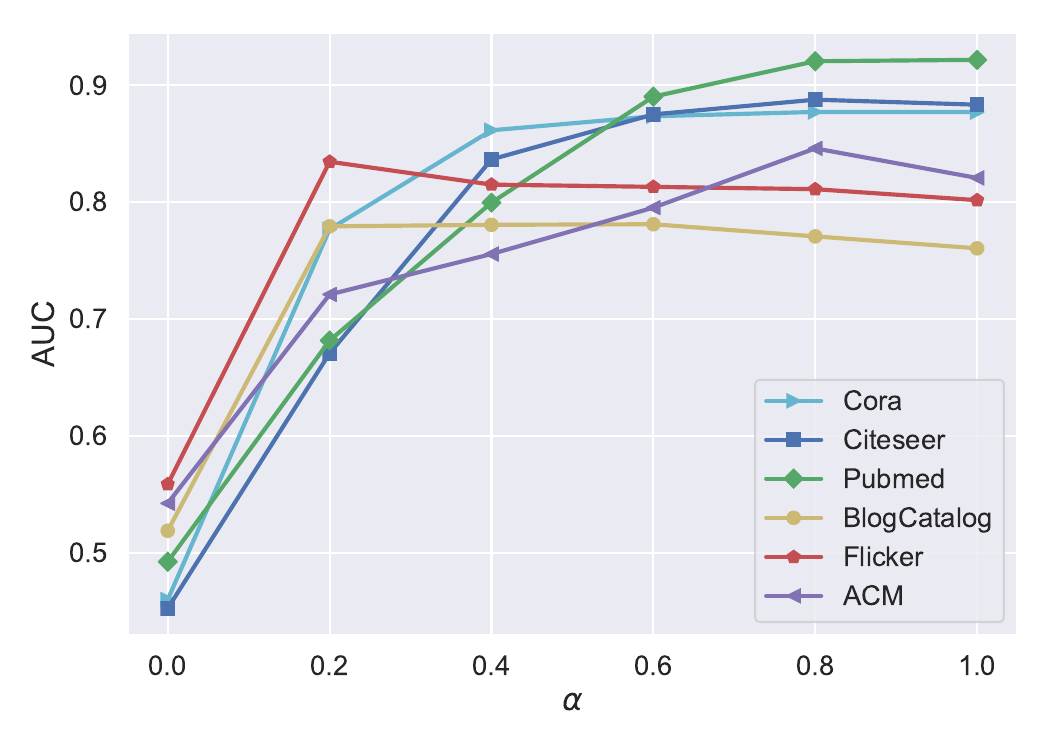}
\end{tabular}
\caption{Effect of hyperparameter $\alpha$ on anomaly detection performance of our model using AUC as evaluation metric.}
\label{Fig:AblationStudy}
\end{figure}

We also analyze the effect of the trade-off hyperparameter $\alpha$ on model performance, and the results are shown in Figure~\ref{Fig:AblationStudy}. When $\alpha=0$, our loss function reduces to the structure reconstruction loss and when $\alpha=1$, it reduces to the feature reconstruction loss. The structure reconstruction loss evaluates the extent to which our model accurately represents the original adjacency matrix, whereas the feature reconstruction loss assesses the fidelity of the reconstructed node feature representation. As can be observed in Figure~\ref{Fig:AblationStudy}, our model generally yields higher AUC values when $\alpha$ is between 0.5 and 0.8. Therefore, the best detection performance is typically achieved by simultaneously considering the reconstruction errors of both graph structure and node features. Notice that in some datasets such as Flicker and BlogCatalog, assigning higher weight to the reconstruction error of the topological structure results in better performance than the reconstruction error of nodal attributes. For the other datasets, giving more weight to the reconstruction error of node features yields better results. This suggests that using node features and graph structure is vital to the model performance in identifying anomalies on data-structured data, allowing for a more accurate and comprehensive understanding of abnormal behavior in graphs. While the graph structure provides a global view of the graph topology, the node features offer a more localized perspective, capturing fine-grained details about individual nodes. Integrating both aspects enables our anomaly detection model to capture anomalies that might be missed by considering only one type of information. In fact, anomalies that exhibit complex patterns, which cannot be solely captured by either structure or features alone, can be better detected when both aspects are taken into account.

\subsection{Ablation Study}
Since there are several components in our proposed model architecture, we analyze their impacts via an ablation study. Specifically, we investigate the effectiveness of the pooling and denoising operations on the anomaly detection performance of our model. To this end, we conduct experiments for ablation studies by removing components individually. The details of our experiments are reported in Table~\ref{Tab:Ablation study} in terms of the AUC metric on six benchmark datasets. As can be seen, the removal of each component leads to a drop in performance. To explore the effect of the denoising operation using spectral graph wavelets, we remove it from our model framework and report the AUC results in Table~\ref{Tab:Ablation study}. This shows that incorporating a denoising operation to our model helps remove the amplified noise introduced during the decoding stage. The denoising operation uses a ReLU function and a trainable parameter matrix to linearly transform the noise, keeping the useful coefficients that are above zero and eliminating the noise that is below zero. This helps improve the model's performance by reducing the effect of the amplified noise on the reconstructed node feature matrix.

On the other hand, the significance of the pooling operation becomes evident when it is excluded from the proposed model architecture. This omission results in a substantial decrease in AUC values, as demonstrated in Table~\ref{Tab:Ablation study}. The effectiveness of our pooling operation in detecting anomalies in graph data is largely attributed to the fact that locality-constrained linear coding focuses on capturing local patterns within the graph, thereby extracting more discriminative features that are specific to the local context, which is often crucial for detecting anomalies. Moreover, localized representation allows our model to better capture the subtle variations and deviations that signify anomalous behavior. In addition, LCPool incorporates the concept of locality, which encourages the extraction of robust features.

\begin{table}[!htb]
\setlength{\tabcolsep}{2.3pt}
\caption{Ablation analysis (AUC ($\%$)) on six datasets. Performance better than the default version is boldfaced.}
\small
\medskip
\centering
\begin{tabular}{lcccccc}
\toprule[1pt]
Method&BlogCatalog & Flicker & ACM & Cora & Citeseer & Pubmed\\
\midrule[.8pt]
w/o denoising & 80.01 & 76.58 & 74.21 & 75.12 & 83.86 & 82.24\\
w/o pooling & 78.79 & 75.87 & 77.10 & 77.22 & 78.56 & 76.23\\
\midrule[.8pt]
Ours & \textbf{82.85} & \textbf{83.12} & \textbf{84.69} & \textbf{88.46} & \textbf{91.34} & \textbf{92.81}\\
\bottomrule[1pt]
\end{tabular}
\label{Tab:Ablation study}
\end{table}

\subsection{Discussions}
The better performance of our proposed method is largely attributed to the combination of an effective graph encoder-decoder architecture, the utilization of LCPool for capturing local patterns, the denoising operation, and the integration of graph structure and node features. The graph encoder-decoder architecture empowers our model to learn more effective and discriminative representations of the graph-structured data. This enables it to better capture the underlying characteristics and structures of the data, aiding in the identification of anomalies. The adoption of LCPool enables our model to focus on local information, leading to the extraction of more robust and representative features. This local focus is crucial for accurately identifying anomalies and distinguishing them from normal graph nodes, especially in scenarios where anomalies exhibit complex and subtle patterns. The denoising operation, facilitated by spectral graph wavelets, mitigates the impact of noise during the decoding stage, resulting in improved reconstruction of the node feature matrix and enhanced anomaly detection performance. Moreover, the integration of both graph structure and node features in the encoding-decoding process provides a more comprehensive view of the graph, allowing our model to capture complex patterns and correlations between nodes and their features. This holistic understanding of the graph data further bolsters the model's anomaly detection capabilities. While our model has demonstrated strong anomaly detection capabilities, there are still some limitations that warrant consideration. For instance, the model's performance may be influenced by the complexity of anomalies present in the data. Different types of anomalies may require specific adaptations or additional mechanisms for improved detection. Also, generalizing the proposed method to completely different domains or highly specialized graph data remains to be fully explored. Overall, our method presents an effective approach to graph-based anomaly detection, benefiting from the interplay of various components that collectively contribute to its superior performance. However, further exploration is necessary to address the identified limitations and fully realize the model's potential in diverse anomaly detection tasks.

\section{Conclusion}
In this paper, we introduced a graph encoder-decoder model for unsupervised anomaly detection. We also proposed a novel pooling strategy that utilizes locality-constrained linear coding for feature encoding. This pooling mechanism involves solving a least-squares optimization problem with a locality regularization term to obtain a cluster assignment matrix. By considering locality, our pooling operation reduces the impact of irrelevant information present in the global graph structure, leading to more robust and representative feature extraction, which is essential for accurately identifying anomalies and distinguishing them from normal graph nodes. In the encoding stage of our model architecture, we used a multi-layer graph convolutional network encoder, followed by the pooling operation. In the decoding, we employed an unpooling operation, followed by a graph deconvolutional network decoder to decode the graph-structured data. Through our experimental evaluations, we demonstrate that our model, which incorporates the proposed pooling and unpooling layers in conjunction with locality-constrained linear coding, outperforms competing baselines on six benchmark datasets across a variety of evaluation metrics, showcasing its superiority in anomaly detection tasks. For future work, we plan to enhance the interpretability and explainability of the model's anomaly detection results in an effort to provide more transparent explanations for the detected anomalies, such as identifying specific graph features or patterns that contribute to the anomaly scores.

\subsection*{Compliance with Ethical Standards}

\smallskip\noindent{\textbf{Conflict of Interest}}\quad The authors declare that they have no financial or personal interests to disclose.

\smallskip\noindent{\textbf{Funding}}\quad This work was supported in part by Natural Sciences and Engineering Research Council of Canada.

\smallskip\noindent{\textbf{Data Availability}}\quad The datasets used in the experiments are publicly available.

\bibliographystyle{ieeetr}
\bibliography{references} 

\end{document}